\documentclass[journal]{IEEEtai}

\usepackage[colorlinks,urlcolor=blue,linkcolor=blue,citecolor=blue]{hyperref}
\usepackage{color,array}
\usepackage{graphicx}
\usepackage{adjustbox}
\usepackage{amsmath,amsfonts}
\usepackage{algorithmic}
\usepackage{array}
\usepackage{multirow}
\usepackage{hyperref}
\usepackage{cellspace}
\usepackage{threeparttable}
\usepackage[caption=false,font=normalsize,labelfont=sf,textfont=sf]{subfig}
\usepackage{textcomp}
\usepackage{stfloats}
\usepackage{url}
\usepackage{verbatim}
\usepackage{graphicx}
\def\BibTeX{{\rm B\kern-.05em{\sc i\kern-.025em b}\kern-.08em
    T\kern-.1667em\lower.7ex\hbox{E}\kern-.125emX}}
\usepackage{balance}

\begin{document}
\title{GFE-Mamba: Mamba-based AD Multi-modal Progression Assessment via Generative Feature Extraction from MCI} 

\author{Zhaojie Fang, Shenghao Zhu, Yifei Chen, Binfeng Zou, Fan Jia, Chang Liu, Xiang Feng, \\ Linwei Qiu, Feiwei Qin, Jin Fan, Changbiao Chu and Changmiao Wang
\thanks{This work was supported by Natural Science Foundation of Zhejiang Province (No. LY21F020015), the Open Project Program of the State Key Laboratory of CAD\&CG (No. A2304), Zhejiang University, GuangDong Basic and Applied Basic Research Foundation (No. 2022A1515110570), Innovation Teams of Youth Innovation in Science and Technology of High Education Institutions of Shandong Province (No. 2021KJ088) and National Undergraduate Training Program for Innovation and Entrepreneurship (No. 202310336074). Data used in the preparation of this article was obtained from the Alzheimer's Disease Neuroimaging Initiative (ADNI) database (adni.loni.usc.edu). The ADNI researchers contributed data but did not participate in analysis or writing of this report. The data was made available at the ADNI database (\href{www.loni.usc.edu}{www.loni.usc.edu}). \textit{(Corresponding author: Feiwei Qin)}}
\thanks{Zhaojie Fang, Shenghao Zhu, Yifei Chen, Binfeng Zou, Fan Jia, Chang Liu, Xiang Feng, Feiwei Qin and Jin Fan are with Hangzhou Dianzi University, Hangzhou, 310018, China (e-mail:(21321206, 22320220, chenyifei, 21321333, 18120401, 22330506, 21070208, qinfeiwei, fanjin)@hdu.edu.cn). Linwei Qiu is with Beihang University, Beijing, 100191, China (e-mail:qiulinwei@buaa.edu.cn). Changbiao Chu is with Xuanwu Hospital, Capital Medical University, Beijing, 100053, China (e-mail:chucb@xwhosp.org). Changmiao Wang is with Shenzhen Research Institute of Big Data, Shenzhen, 518172, China (e-mail:cmwangalbert@gmail.com).}

}
\markboth{Journal of IEEE Transactions on Artificial Intelligence, Vol. 00, No. 0, December 2024}
{First A. Author \MakeLowercase{\textit{et al.}}: Bare Demo of IEEEtai.cls for IEEE Journals of IEEE Transactions on Artificial Intelligence}

\maketitle

\begin{abstract}
Alzheimer's Disease (AD) is a progressive, irreversible neurodegenerative disorder that often originates from Mild Cognitive Impairment (MCI). This progression results in significant memory loss and severely affects patients' quality of life. Clinical trials have consistently shown that early and targeted interventions for individuals with MCI may slow or even prevent the advancement of AD. Research indicates that accurate medical classification requires diverse multimodal data, including detailed assessment scales and neuroimaging techniques like Magnetic Resonance Imaging (MRI) and Positron Emission Tomography (PET). However, simultaneously collecting the aforementioned three modalities for training presents substantial challenges. To tackle these difficulties, we propose GFE-Mamba, a multimodal classifier founded on Generative Feature Extractor. The intermediate features provided by this Extractor can compensate for the shortcomings of PET and achieve profound multimodal fusion in the classifier. The Mamba block, as the backbone of the classifier, enables it to efficiently extract information from long-sequence scale information. Pixel-level Bi-cross Attention supplements pixel-level information from MRI and PET. We provide our rationale for developing this cross-temporal progression prediction dataset and the pre-trained Extractor weights. Our experimental findings reveal that the GFE-Mamba model effectively predicts the progression from MCI to AD and surpasses several leading methods in the field. Our source code is available at \href{https://github.com/Tinysqua/GFE-Mamba}{https://github.com/Tinysqua/GFE-Mamba}.
\end{abstract}


\section{Introduction}

\IEEEPARstart{A}{lzheimer's} Disease (AD) is a prevalent neurodegenerative condition in older adults, affecting memory, cognitive abilities, and daily life activities \cite{c40alzheimer20182018}. It often develops from Mild Cognitive Impairment (MCI), especially amnestic MCI (aMCI), which is primarily marked by memory issues. Although people with aMCI experience significant memory loss, their cognitive decline hasn't reached the stage of dementia. Predicting whether aMCI patients will progress to AD within one to three years is essential for prognosis. Early identification of high-risk individuals allows for tailored treatment and intervention plans, which can slow disease progression and improve quality of life \cite{c41crous2017alzheimer}. Additionally, early prediction helps patients and their families make informed decisions and prepare both psychologically and practically. Research supports the idea that early detection and targeted interventions can significantly decelerate or even stop the progression of AD. Prognostic predictions assist physicians in adopting appropriate management and treatment strategies \cite{c42jo2019deep}. For patients at high risk, more intensive interventions, such as medication and cognitive training, are often implemented. Medications, including cholinesterase inhibitors like donepezil and NMDA receptor antagonists like memantine, can alleviate cognitive symptoms and delay disease progression. For patients not expected to decline soon, regular monitoring and lifestyle interventions are recommended. Routine cognitive assessments and annual neuroimaging can catch potential changes early, while non-drug approaches like cognitive training can help maintain or enhance cognitive functions. Lifestyle changes, such as better diet, exercise, and psychological support, can improve overall health and resilience against diseases \cite{c7livingston2020dementia}.

\begin{figure*}
    \includegraphics[width=\textwidth]{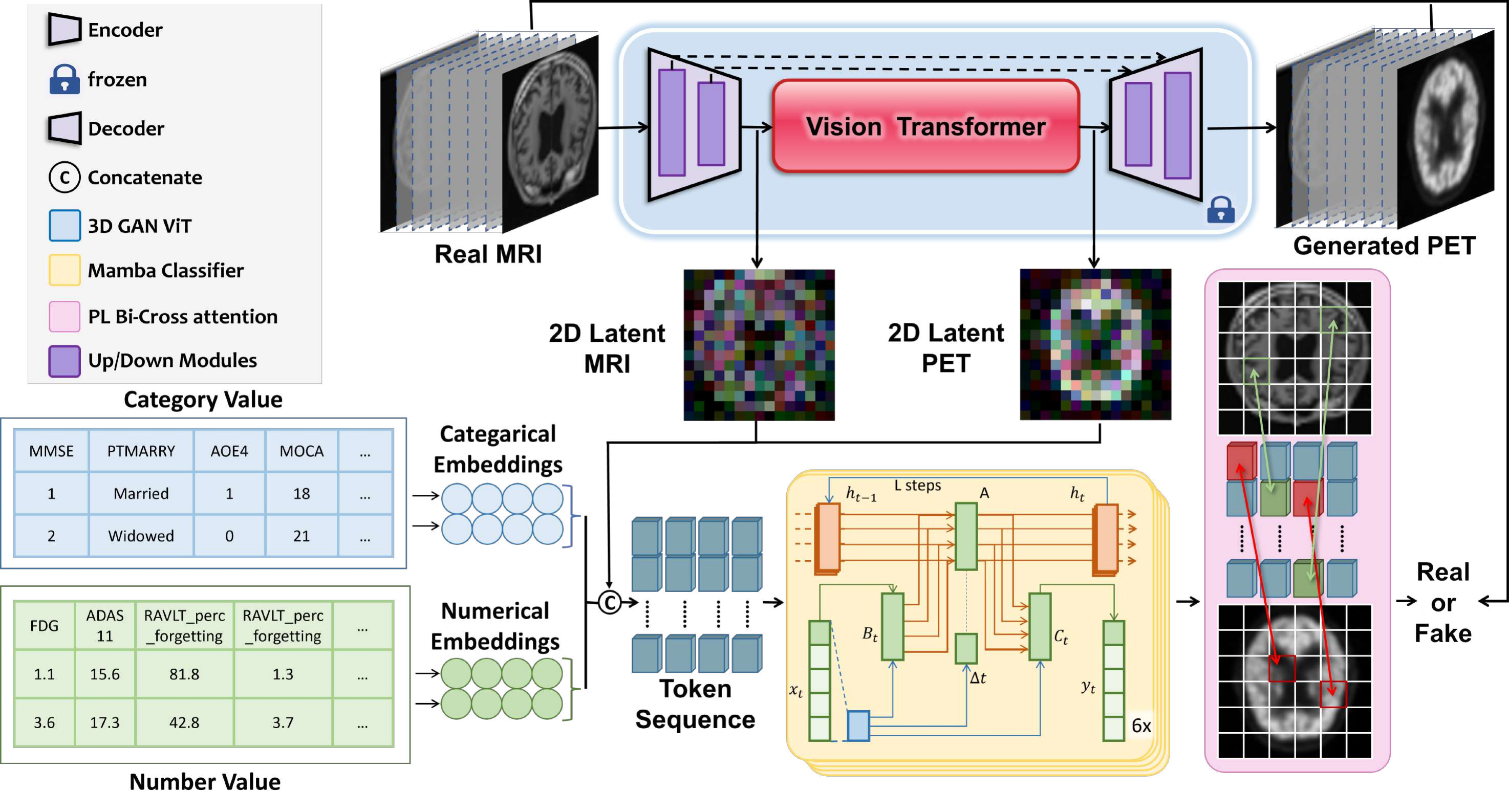}
    \caption{The overall architecture of GFE-Mamba. It contains a 3D GAN-ViT that was pre-trained on the MRI to PET generation task. 2D Latent MRI and PET features extracted from the 3D GAN-ViT are fused with the scale information and then both are fed into the Multimodal Mamba classifier. The output of the classifier and Pixel-level MRI/PET will predict a binary classification outcome after Pixel-Level Bi-Cross Attention.
}
    \label{fig1}
\end{figure*}

Currently, prognostic predictions for patients with aMCI rely on neuroimaging, cognitive assessments, and biomarker testing. The Knowledge Scale is frequently used in initial diagnostics to screen for aMCI, though its accuracy may be affected by individual variations, making it mainly suitable for preliminary evaluations \cite{c1jia2014prevalence}. MRI provides detailed images of brain structures, facilitating the observation of changes such as brain volume reduction and cortical atrophy. In contrast, Positron Emission Tomography (PET) offers insights into brain metabolic activity and $\beta$-amyloid accumulation, which are essential markers for early detection of AD \cite{c2jack2018nia}. Despite its advantages, PET imaging is time-intensive, expensive, and technically challenging. Its high sensitivity and specificity for detecting subtle changes in brain metabolism and early AD markers, such as $\beta$-amyloid, make it a potent tool for assessing AD risk and progression. However, the complexity of PET imaging, which involves specific radiotracers, precision detectors, and advanced image reconstruction techniques, adds to its cost and difficulty. The necessity for skilled personnel further limits its widespread use. Nonetheless, PET imaging's ability to provide crucial predictive information and monitor AD progression remains invaluable. Effective prediction of aMCI progression to AD requires considering multiple risk factors and integrating various diagnostic tools to achieve a comprehensive assessment. This multidisciplinary approach enhances the accuracy of prognostic predictions and aids in the timely implementation of personalized interventions.

Despite the numerous methods currently employed to predict the progression from aMCI to AD, significant challenges and limitations persist, particularly concerning the accuracy and reliability of these predictions \cite{c9mattsson2020longitudinal}. This study seeks to improve prognostic predictions by synthesizing multiple methods, exploring more effective approaches, and offering valuable insights for enhanced patient management and therapeutic outcomes \cite{c11cummings2019role}. To address these challenges, we propose the AD prediction model GFE-Mamba, which automates the classification and prediction of AD using MRI. This model integrates several advanced techniques, including a 3D GAN-ViT, a Vision Transformer (ViT) bottleneck layer, a mamba block backbone network, and Pixel-Level Bi-Cross Attention. These components work together to efficiently extract pathological features from MRI, with the Generative Feature Extraction (GFE) module synthesizing PET. By incorporating scale information into the mamba block backbone network, our approach significantly enhances the precision of AD classification. This comprehensive method aims to provide a more reliable and practical diagnostic tool for the early detection and management of patients at risk of transitioning from aMCI to AD. The primary contributions of this paper are as follows: 

1) \textbf{3D GAN-ViT From MRI To PET}: We employ a 3D GAN as the backbone, integrated with a ViT for generative task learning. This combination facilitates GFE for the Mamba Classifier, effectively capturing spatial features from both MRI and PET images.

2) \textbf{Multimodal Mamba Classifier}: We propose a Mamba Classifier that processes extensive scale information and 3D images. This classifier integrates sequences through six mamba blocks, followed by averaging and linear layers, to produce the final classification.
    
3) \textbf{Pixel-Level Bi-Cross Attention}: We implement a pixel-level cross-attention strategy to improve the classifier’s ability to capture often-overlooked pixel-space information from both MRI and PET images efficiently. 

4) \textbf{Dataset Construction}: To demonstrate the generalizability of our approach, we constructed three datasets using data from the Alzheimer's Disease Neuroimaging Initiative (ADNI). The first dataset consists of paired MRI and PET (MRI-PET dataset). The following datasets, focused on one-year and three-year MCI-AD progression, are utilized to train the classifier. We provide a comprehensive explanation of how these datasets were constructed and will make the dataset acquisition and processing code available to the public on GitHub at \href{https://github.com/Tinysqua/GFE-Mamba}{https://github.com/Tinysqua/GFE-Mamba}.
\begin{figure*}
    \includegraphics[width=\textwidth]{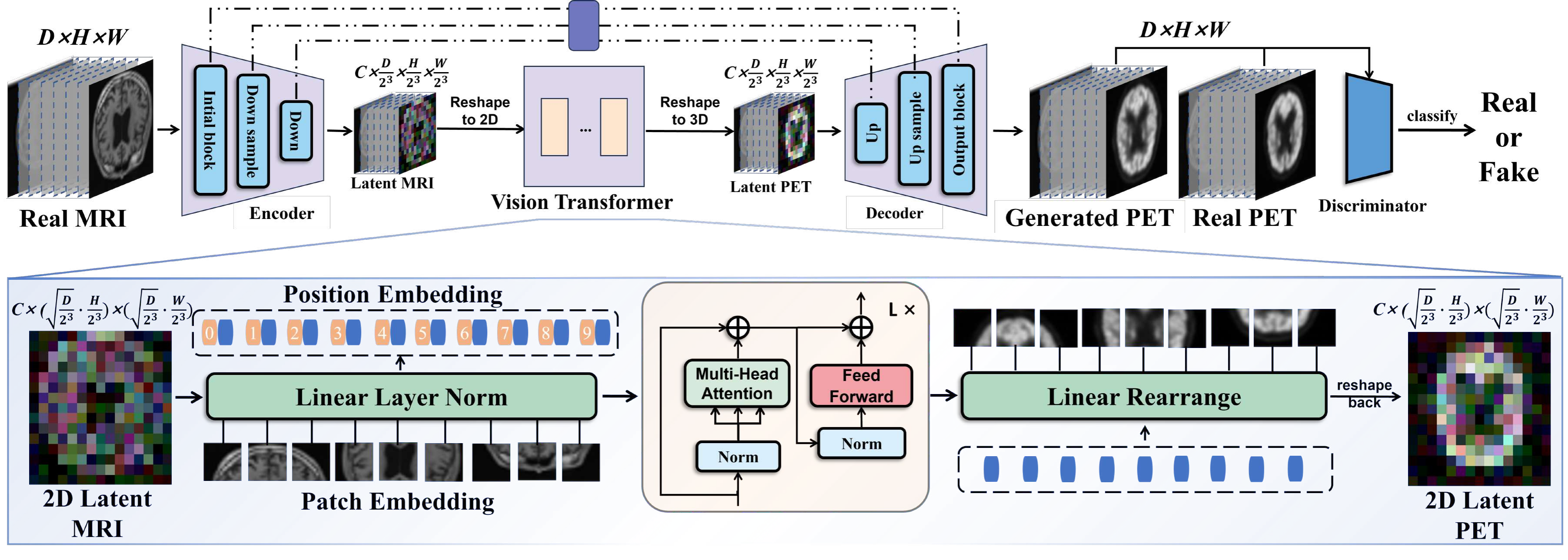}
    \caption{The architecture of 3D GAN-ViT. The MRI data with a shape of \( D \times H \times W \) is compressed through an encoder to obtain a 2D latent MRI representation. After dividing it into patches, it is fed into a Vision Transformer (ViT). The output is then reshaped into a latent PET representation, which is subsequently passed through a decoder to reconstruct the PET, where the generated PET will be sent together with the real PET into the Discriminator to assess the quality of generation and training.
}
    \label{fig2}
\end{figure*}

\section{Related Work}


\subsection{Traditional and Machine Learning Prediction Methods}

Traditional methods for predicting the progression of AD heavily rely on cognitive assessments and biomarker testing. Tools such as the Mini-Mental State Examination \cite{c33folstein1975mini} and the Montreal Cognitive Assessment \cite{c34nasreddine2005montreal} have been developed to evaluate cognitive disorders and screen for dementia. Biomarker testing, pioneered by Leland Clark Jr. \cite{c35clark1962electrode}, measures AD-related pathological changes by assessing beta-amyloid and tau proteins in the cerebrospinal fluid. Despite these advancements, predicting the transition from aMCI to AD remains challenging. These limitations highlight the need for more accurate methods for early prediction of AD progression.

With technological advancements, traditional methods for predicting Alzheimer’s disease have evolved to incorporate machine learning techniques, significantly improving prediction accuracy. Escudero \textit{et al.} \cite{c12escudero2011early} utilized multimodal data, including clinical, neuroimaging, and biochemical information, and applied k-means clustering to categorize subjects into pathological and non-pathological groups. They also employed regularized logistic regression for classification \cite{c13misra2009baseline, c14ye2012sparse}. Wan \textit{et al.} \cite{c15wan2012sparse} proposed a sparse Bayesian multi-task learning algorithm to enhance computational efficiency. Building on this foundation, Young \textit{et al.} \cite{c16young2013accurate} achieved high prediction accuracy on the ADNI database using the Gaussian Process classification algorithm, which integrated multimodal data through a hybrid kernel function.  These studies collectively highlight the potential of machine learning in enhancing the accuracy and reliability of AD predictions.

\subsection{Neural Network Based Prediction Method}

With the increase in computer processing power and the development of deep neural network technologies, research into AD prediction methods has gained significant momentum \cite{elazab2024alzheimer, yuan2019prostate}. Liu \textit{et al.} \cite{c21liu2023monte} leveraged CNNs to extract image features from brain regions associated with cognitive decline, which were then combined with non-image data using a SVM classifier. Qiu \textit{et al.} \cite{c22qiu2020development} used a Fully Convolutional Network to create high-resolution disease probability maps from MRI, integrating features from high-risk regions with non-imaging data to classify AD. Further advancing the field, Liu \textit{et al.} \cite{c23liu2023cascaded} introduced the 3MT architecture, which integrates multimodal information through cross-attention and incorporates a modal dropout mechanism. Rahim \textit{et al.} \cite{c24rahim2023prediction} proposed a hybrid framework that combines a 3D CNN with a bidirectional RNN. El-Sappagh \textit{et al.} \cite{c25el2020multimodal} combined a stacked CNN with a Bidirectional Long Short-Term Memory (BiLSTM) network to predict AD progression, achieving an accuracy of 92.62\% by fusing five types of time-series multimodal data. Building on these advancements, Wang \textit{et al.} \cite{c26wang2023hypergraph} developed a multimodal learning framework that employs hypergraph regularization through a graph diffusion approach, achieving an accuracy of 96.48\%. These developments underscore the growing potential of integrating various neural network architectures and multimodal information to enhance the accuracy of AD prediction models.

\begin{figure*}
    \includegraphics[width=\textwidth]{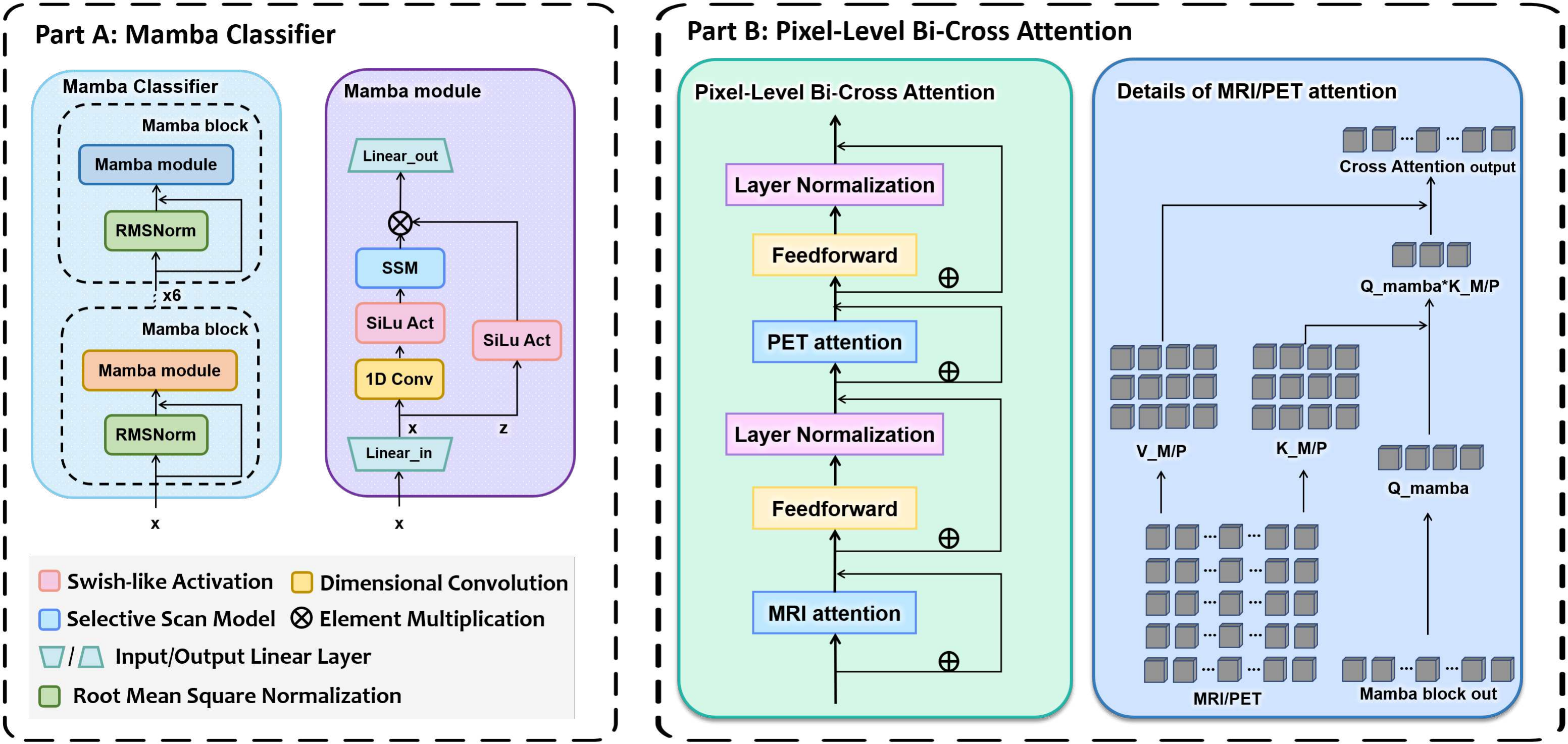}
    \caption{The framework and component modules of the Mamba Classifier and Pixel-Level Bi-Cross Attention. Part A is details of the Mamba Classifier and the Mamba module in it. Part B is details of the Pixel-Level Bi-Cross Attention module.}
    \label{fig3}
\end{figure*}

Recent advancements in related fields offer valuable insights into classification and recognition techniques. Zha \textit{et al.} \cite{c38zeng2018improving} showed that integrating global context with local object features substantially improves remote sensing scene classification. In a similar vein, Xu \textit{et al.} \cite{c39ge2020occluded} developed an identity-diversity inpainting technique for recognizing occluded faces, thereby enhancing recognition accuracy in real-world scenarios. Pan \textit{et al.} \cite{c43pan2024decgan} proposed a novel decoupling GAN for detecting abnormal neural circuits in AD, thereby improving the accuracy and robustness of AD diagnosis in clinical settings. Shi \textit{et al.} \cite{c45shi2022asmfs} proposed an Adaptive-Similarity-based Multi-modality Feature Selection method for classifying and its prodromal stage MCI. Park \textit{et al.} \cite{c44park2023prospective} proposed a prospective classification method for predicting the conversion of AD from MCI. These studies highlight the advantages of merging diverse feature sets with advanced methodologies to refine classification and recognition across various applications.

\subsection{Comparison with Existing Work}

Currently, various methods utilize deep learning architectures, particularly 3D-CNNs with residual connections, to predict AD progression using MRI or PET scans. These approaches allow for simultaneous predictions by incorporating both MRI/PET scans and scale information, demonstrating that multimodal data can enhance prediction accuracy. However, many existing methods rely heavily on large amounts of paired data and incur high training costs to align multiple modalities, significantly limiting their clinical applicability. To overcome these challenges, our approach separates the alignment of multimodal data into two distinct tasks: generation and classification. This separation reduces the reliance on paired data and simplifies the training process for multimodal classification.

\section{Methods}
Our approach, depicted in Fig. \ref{fig1}, comprises three main components: the MRI to PET Generation Network, the Multimodal Mamba Classifier, and the Pixel-Level Bi-Cross Attention mechanism. The MRI to PET Generation Network is first trained using a comprehensive dataset of paired MRI and PET images. This training enables the network to generate PET from MRI scans when PET is unavailable. By extracting information from MRI, the network creates PET features, which are then passed to the classifier for multimodal fusion. The Multimodal Mamba Classifier efficiently processes the fused data, including both tabular and intermediate image features, to make accurate classification decisions. To enhance this process, the Pixel-Level Bi-Cross Attention mechanism works at the pixel level, integrating MRI and PET to overcome the classifier's limitations in handling shallow spatial image information. This integrated approach effectively combines additional PET information, even when only MRI data is present, allowing for more precise predictions of a patient's likelihood of progressing from MCI to AD in the future.

\subsection{3D GAN-ViT on MRI-PET  Task}
MRI and PET scans provide essential structural and functional information about the brain, making them vital for predicting the progression of AD. However, incorporating both types of data into a classification network in clinical settings poses two major challenges. The first challenge is the need for paired MRI, PET, and labeled data, which is difficult to obtain, resulting in a limited training dataset that can easily lead to model overfitting. The second challenge arises in real clinical scenarios, where patients often undergo only the more affordable MRI scans and a few standard tests, leading to a lack of multimodal data. Despite these challenges, paired MRI and PET are still available. 

\begin{figure*}
\centering
    \includegraphics[width=\textwidth]{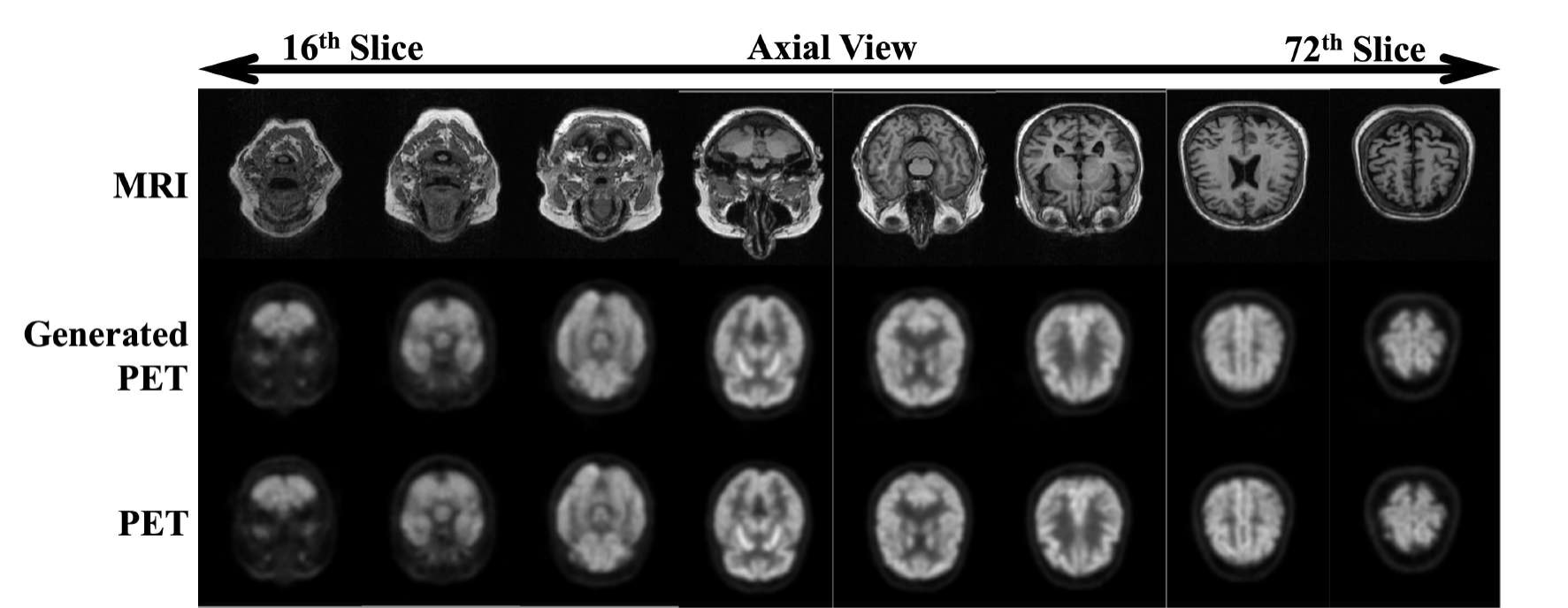}
    \caption{The results of 3D GAN-ViT. From left to right, these are slices of 3D images in the axial direction. To eliminate the black background around the images, the first column on the left starts with the 16th slice, with each subsequent column adding 8 slices. Within each column, from top to bottom, are the MRI, Generated PET, and PET images.
}
    \label{fig11}
\end{figure*}

To address these issues, we propose developing a generation network that translates MRI data into PET representations. This approach facilitates effective representation learning using paired data and allows for the network's pre-training process to be completed in advance. For this generation network, we utilize a 3D GAN-ViT architecture. Specifically, we use the 3D GAN network as the foundation, replacing the original ResNet middle block with a ViT. This modification enhances the network's ability to generate accurate PET representations from MRI data. In Fig. \ref{fig11}, the generated PET and the real one are shown together. These slices visually demonstrate the generation effect of 3D GAN-ViT, which is almost identical to the real one. Therefore, even without paired data, the intermediate features from 3D GAN-ViT can transfer the information of an MRI corresponding to PET.

\textbf{3D GAN as Generative Feature Extractor.} Ian \textit{et al.} introduced a Generative Adversarial Network (GAN) \cite{c1jia2014prevalence} designed to produce high-quality images, which can be effectively used in various tasks through representative learning. Building on this, the 3D GAN \cite{c2jack2018nia} model extends the original GAN framework to three-dimensional medical imaging applications. We adopt this 3D GAN as the core of our generation network, which consists of two main components: a Discriminator ($D$) and a Generator ($G$). The Discriminator guides the Generator to create realistic PET images by learning to extract and transform features from MRI scans into PET features using extensive datasets. This capability ensures that features from both modalities can be leveraged, even when only MRI data is available.

The architecture of our 3D GAN-ViT network, depicted in Fig. \ref{fig2}, includes an Encoder/Decoder module with convolutional layers as the base and ViT layers integrated in the middle. The Encoder comprises three down-sampling modules, each consisting of a max pooling layer, group normalization layer, convolutional layer, and ReLU activation layer, with channel sizes of 64, 128, and 256, respectively. Similarly, the Decoder features three up-sampling modules with channel sizes of 256, 128, and 64, each including a group normalization layer, transpose convolutional layer, and ReLU activation layer, mirroring the down-sampling structure. In this setup, the MRI input ($\mathbf{x}_M$) is processed by the Generator through the encoder and decoder to produce the PET output ($\mathbf{y}_P$), which then serves as input for the Discriminator. The Discriminator uses the same three down-sampling modules to analyze the input PET ($\mathbf{X}_{PET}$), resulting in a feature map ($\mathbf{Y}$) that contributes to the computation of the loss function.

The 3D GAN network's loss function comprises two parts: Generator loss and Discriminator loss. The Generator loss is expressed as:
\begin{equation}
    \begin{aligned}
    \mathcal{L}(G) = & \sum_{x_M \in X_M, y_P \in Y_P} \left\Vert G(\mathbf{x}_M) - \mathbf{y}_P \right\Vert^2 \\
    & + \log(1 - D(G(\mathbf{x}_M))) + \left\Vert VGG(\mathbf{x}_M) - VGG(\mathbf{y}_P) \right\Vert^2
    \end{aligned}
\end{equation}
where the first term is the Mean Squared Error reconstruction loss between real and generated PET images, the second term is the adversarial loss of the generator, and the third term is the perceptual loss obtained through VGG19 \cite{c6nebel2018understanding}.

The Discriminator loss is defined as:
\begin{equation}
    \mathcal{L}(D)=\sum_{x_M\in X_M,y_P\in Y_P}\log(1-D(\mathbf{y}_P))+\log(D(G(\mathbf{x}_M))),
\end{equation}
where the first term represents the adversarial loss on real PET images, and the second term accounts for the adversarial loss on the generated PET images.

\textbf{Vision Transformer as Middle Block.}
The Encoder and Decoder of the 3D GAN-ViT network are pivotal in transforming MRI data into PET by compressing it into a latent space and then reconstructing it. To optimize this transformation, we replace the traditional ResNet middle block of the 3D GAN with a ViT module. This modification is vital because the classifier's backbone processes sequences, and integrating spatial features directly could lead to a loss of spatial information. The ViT resolves this by applying inter-attention to the image's flattened vectors in the hidden space.

Initially, the Encoder compresses the MRI data into a latent space, denoted as \( \mathbf{x}_{LM} \in \mathbb{R}^{H \times W \times D \times C} \). This 3D feature map is then flattened into a 2D feature map, resulting in \( \mathbf{x}_{LM} \in \mathbb{R}^{(H \cdot \sqrt{D}) \times (W \cdot \sqrt{D}) \times C} \), which is subsequently processed by the ViT. The feature map is divided into a series of image patch sequences \( \mathbf{x}_{LMP} \in \mathbb{R}^{N \times (p^2 \cdot C)} \) through Patch Embedding, where \( p \) is the patch size and \( N \) equals \( \frac{(H \cdot \sqrt{D})}{p} \).

Once the 3D feature map is converted into a sequence, it passes through the transformer encoder, comprising four transformer blocks. After processing, the sequence \( \mathbf{x}_{LMP} \in \mathbb{R}^{N \times (p^2 \cdot C)} \) is resized back to \( \mathbf{x}_{LP} \in \mathbb{R}^{H \times W \times D \times C} \) for the Decoder to generate the PET image. Following pre-training, the latent representations \( \mathbf{x}_{LMP} \) for MRI and \( \mathbf{x}_{LPP} \) for PET effectively encapsulate information from both modalities. These comprehensive representations are subsequently compiled and provided directly to the classifier for the next stage of information fusion.

\subsection{Multimodal Mamba Classifier}
\textbf{Time Interval Extraction.}
To effectively predict the progression from MCI to AD, it is essential to establish a specific prediction time frame, such as determining if a patient with MCI will transition to AD within 180 days. If a 180-day period is set, the model aims to predict whether a patient with MCI will develop AD after this duration. Thus, the training dataset must represent this interval between diagnoses accurately. However, assembling such a dataset is challenging because it is difficult to ensure that the time between diagnoses for each patient is precisely 180 days. To overcome this challenge, we employ a dynamic strategy. We record the actual time intervals between diagnoses for each patient and integrate this information into the model's training data, along with assessment scale category values. During inference, we use the average time interval from the training set to make predictions. This method accounts for variations in the timing of patient diagnoses, ensuring that the model effectively predicts the progression from MCI to AD within the specified time frame.

\begin{figure*}
\centering
    \includegraphics[width=\textwidth]{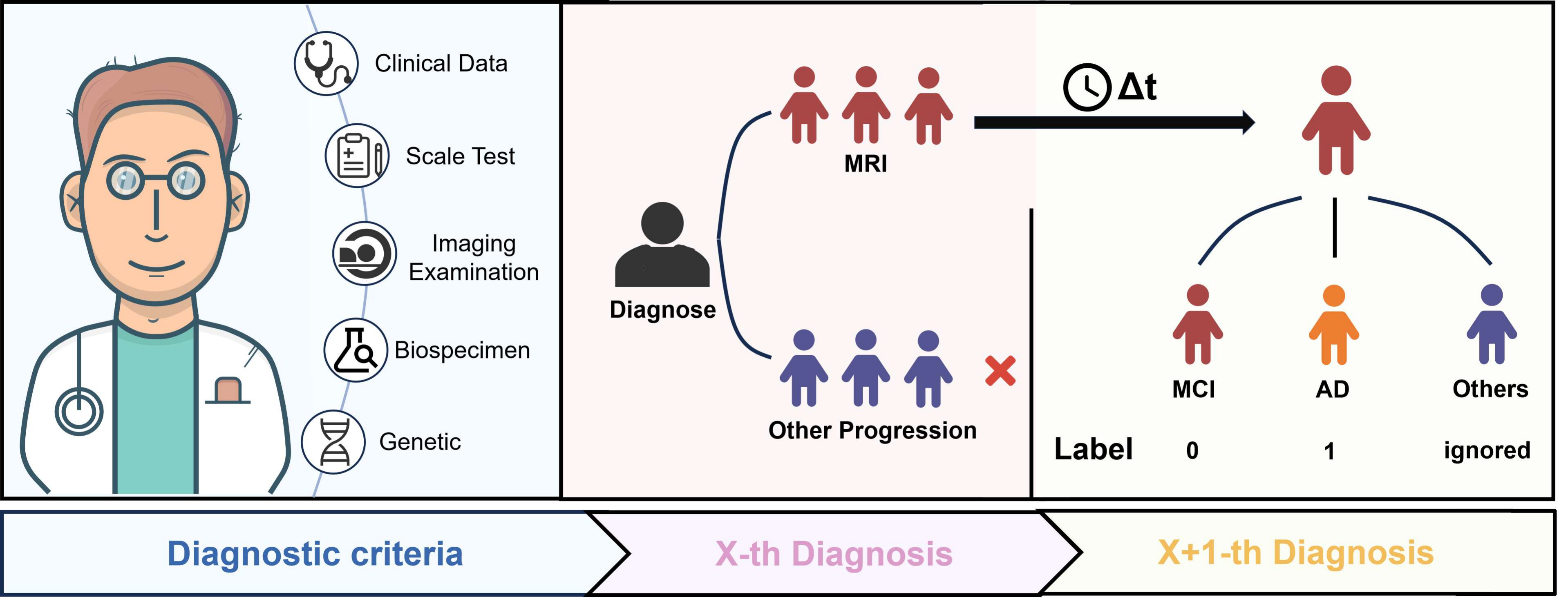}
    \caption{The process of constructing MCI-AD dataset from ADNI dataset. When constructing data, the MRI diagnosis is identified at time x, and then at time x+1, the AD/MCI diagnoses are labeled as positive/negative samples respectively.
}
    \label{fig4}
\end{figure*}

\begin{table}
\centering
\caption{MVoPS/MVoNS: Mean Value of Positive/Negative Samples, SDoPS/SDoNS: Standard Deviation of Positive/Negative Samples, PS/NS: the ratio of Positive/Negative Samples.}
\label{tab1}
\renewcommand{\arraystretch}{1.5}
\begin{adjustbox}{width=0.45\textwidth, center}
\begin{tabular}{l|lllll} 
\hline
Dataset                            & MvoPS & MVoNS & SDoPS & SDoNS & PS/NS    \\ 
\hline
$150<\Delta t<365$  & 6.7   & 7.26  & 1.65  & 2.32  & 134/168  \\
$150<\Delta t<1095$ & 9.06  & 9.57  & 5.21  & 6.27  & 163/218\\
\hline
\end{tabular}
\end{adjustbox}
\end{table}

\begin{figure}
\centering
    \includegraphics[width=\linewidth]{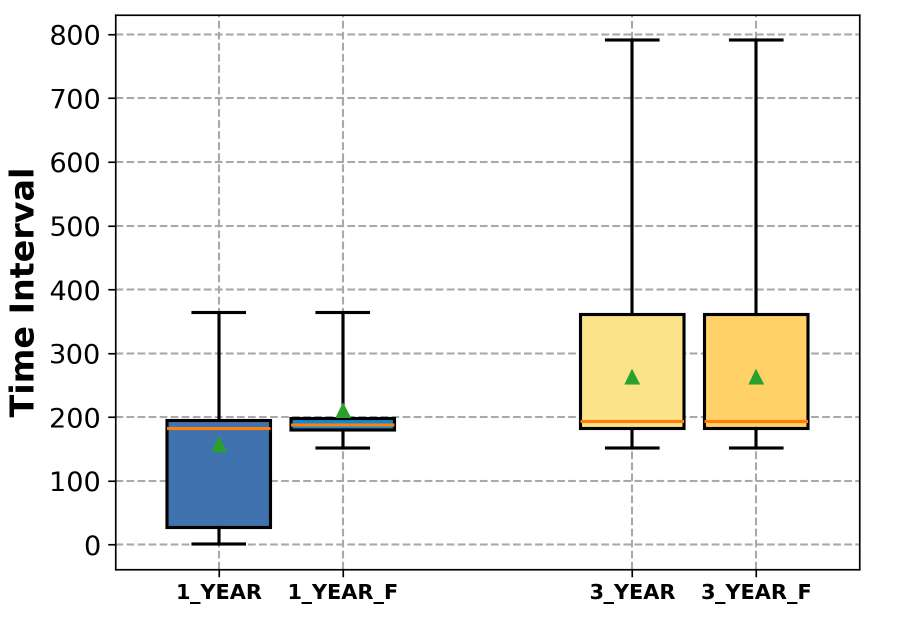}
    \caption{Analysis of the time interval distribution of the one-year and three-year progression dataset. 1 YEAR and 3 YEAR represent one-year and three-year progression, respectively. F indicates dataset that extreme values have been excluded.}
    \label{fig7}
\end{figure}

\textbf{Preprocessing of Assessment Scales.}
To improve the predictive accuracy of our model, we incorporate assessment scales, similar to how physicians use MRI and PET images alongside diagnostic scales. Integrating these scales into the model serves two main purposes. First, the scales offer direct diagnostic assistance. Second, the structured nature of tabular data makes it less noisy compared to images. For effective classification through multimodal fusion, it is crucial to process the assessment scale information and integrate it with image data. We start by categorizing the scale information into discrete category values and continuous numerical values.

\textbf{For the discrete category values:} We convert them into unique heat codes to prevent duplication across different rows. This is achieved by increasing the value in each subsequent column by the maximum number of categories from all previous columns, as expressed by the formula: \(\mathbf{e}_i^{cat}=\mathbf{e}_i^{cat}+\sum_{j=1}^{i-1}\max(\mathbf{e}_j^{cat})\). These transformed values can then be embedded using a linear transformation:
\begin{equation}
    \mathbf{T}_i^{cat}=\mathbf{W}_i^{cat}\mathbf{e}_i^{cat}+\mathbf{b}_i^{cat},
\end{equation}
where \( \mathbf{b}_i^{cat} \) represents the bias for the \( i \)-th feature, and \( \mathbf{W}_i^{cat} \) is the look-up table for the \( i \)-th category \cite{c3petersen2013mild}.

\textbf{For the continuous number values:} The process begins with calculating the mean (\( \mu_i^{num} \)) and standard deviation (\( \sigma_i^{num} \)) for each column. The values are then normalized using the following formula: \( \mathbf{x}_i^{num} = (\mathbf{x}_i^{num} - \mu_i^{num}) / \sigma_i^{num} \). These normalized values are then embedded using a linear transformation:
\begin{equation}
    \mathbf{T}_i^{num}=\mathbf{W}_i^{num}\mathbf{x}_i^{num}+\mathbf{b}_i^{num}.
\end{equation}

After processing both categorical and numerical values, this tabular information is combined with image features. This combination is represented by:
\begin{gather}
    \mathbf{x}=\text{stack}[\mathbf{x}_{LMP},\mathbf{x}_{LPP},\mathbf{T}]\in \mathbb{R}^{(m+n+2N)\times d},
\end{gather}
where \( n \) and \( m \) indicate the number of categorical and numerical rows, respectively. The terms \( x_{\text{LMP}} \) and \( x_{\text{LPP}} \) refer to the features of MRI and PET images within the hidden space of the generative network, and \( d \) signifies the embedding size. Once this processing is complete, \( x \) is input into the classification network, where it is integrated with image data for prediction.

\begin{table*}[]
\begin{center}
\caption{Comparison of GFE-Mamba with other state-of-the-art models in 1-year Dataset and 3-year Dataset (\%).}
\label{tab2}
\begin{adjustbox}{width=\textwidth, center} 
\begin{tabular}{@{}c|ccccc|ccccc@{}}
\hline
\multirow{2}{*}{Method}                                                 & \multicolumn{5}{c|}{1-Year Dataset}                        & \multicolumn{5}{c}{3-Year Dataset}                        \\
\cline {2-11}
& Precision($\uparrow$) & Recall($\uparrow$)    & F1-score($\uparrow$)  & Accuracy($\uparrow$)  & MCC($\uparrow$)       & Precision($\uparrow$) & Recall($\uparrow$)    & F1-score($\uparrow$)  & Accuracy($\uparrow$)  & MCC($\uparrow$)      \\
\hline
ResNet50 \cite{c27he2016deep}$^{\dagger}$ & 81.03 & 78.95 & 79.97 & 73.17 & 59.00 & 77.59 & 75.00 & 76.27 & 69.77 & 52.42 \\
ResNet101 \cite{c27he2016deep}$^{\dagger}$ & 79.31 & 47.37 & 59.31 & 60.00 & 50.57 & 77.14 & 73.33 & 75.18 & 73.33 & 53.33 \\
PENet \cite{huang2020penet}$^{\dagger}$ & 76.92 & 71.42 & 74.07 & 78.125 & 55.31 & 76.19 & 59.25 & 66.66 & 75.00 & 48.11 \\
JSRL \cite{liu2022assessing}$^{\dagger}$ & 89.28 & 92.59 & 90.90 & 92.18 & 84.10 & 86.20 & 92.59 & 89.28 & 90.62 & 81.13 \\
TabTransformer \cite{c28huang2020tabtransformer}$^{\ddagger}$ & 84.29 & 90.57 & 87.32 & 83.58 & 70.22 & 82.86 & 63.33 & 71.79 & 76.00 & 66.64 \\
FTTransformer \cite{gorishniy2021revisiting}$^{\ddagger}$ & 85.71 & 82.75 & 84.21 & 85.93 & 71.57 & 79.31 & 82.14 & 80.70 & 82.81 & 65.25 \\
XGBoost \cite{c29kavitha2022early}$^{\ddagger}$ & 88.14 & 86.53 & 86.92 & 87.37 & 73.90 & 90.0 & 89.95 & 89.97 & 89.58 & 79.53 \\
Qiu et al's \cite{c30qiu2020development}$^{\star}$ & 86.67 & 81.82 & 84.17 & 81.82 & 71.29 & 82.35 & 84.62 & 83.46 & 78.57 & 64.17 \\
Radfusion \cite{zhou2021radfusion}$^{\star}$ & 87.99 & 84.61 & 86.27 & 89.06 & 77.23 & 83.33 & 76.92 & 80.00 & 84.37 & 67.35 \\
Fusion model \cite{c31qiu2022multimodal}$^{\star}$ & 89.83 & 88.21 & 88.91 & 89.87 & 78.06 & 91.43 & 91.25 & 91.34 & 91.25 & 82.50 \\
Zhang et al's \cite{c32zhang2023alzheimer}$^{\star}$ & 76.67 & 83.33 & 79.86 & 58.82 & 48.42 & 70.59 & 60.87 & 65.37 & 73.68 & 48.79 \\
\textbf{GFE-Mamba (Ours)} & \textbf{95.71} & \textbf{96.55} & \textbf{96.13} & \textbf{94.92} & \textbf{91.25} & \textbf{94.83} & \textbf{94.74} & \textbf{94.78} & \textbf{92.31} & \textbf{88.48} \\
\hline
\end{tabular}
\end{adjustbox}
\footnotesize
\begin{flushleft}
$^{\dagger}$: Unimodal Methods of CT. $^{\ddagger}$: Single Modal of Assessment Scale. $^{\star}$: Multimodal Methods.
\end{flushleft}
\end{center}
\end{table*}

\textbf{Mamba Classifier.}
The input \( \mathbf{x} \in \mathbb{R}^{(m+n+2N) \times d} \), which contains a variety of scale information and an extended sequence length due to 3D image features, poses efficiency challenges when using a traditional transformer with quadratic attention complexity for training. To overcome these challenges associated with modeling long sequences, we utilize the Mamba Model \cite{c4hampel2018blood}. After integrating the tabular and image information, this sequence is processed by a classifier composed of six Mamba blocks. The structure of these Mamba blocks is depicted in Fig. \ref{fig3} (Part A).

Each Mamba block begins with RMS Normalization, which normalizes the input sequence by computing the root mean square value of the input activations. This step is crucial for preventing gradient explosion in deep networks. Following normalization, the Mamba module processes the input sequence, and the resulting output is combined with the input residuals, as expressed in the equation:
\begin{equation}
\mathbf{x}_{i+1} = \text{Mamba}(\text{RMSNorm}(\mathbf{x}_i)) + \mathbf{x}_i. 
\end{equation} 
Initially, the input features undergo a linear transformation and are then split into two components: \( \mathbf{x} \) and \( \mathbf{z} \). These components are obtained via the operation \( \mathbf{x}, \mathbf{z} = \text{split}(\text{linear}(\mathbf{x})) \). The \( \mathbf{x} \) segment is processed through a 1D convolution, followed by activation and further processing via the Selective Scan Model (SSM):
\begin{equation}
    \mathbf{y} = \text{SSM}(\text{Conv}(\mathbf{x})).
\end{equation} 

Concurrently, the \( \mathbf{z} \) segment acts as a gating vector, which is element-wise multiplied with the activated \( \mathbf{y} \). Once processed by the Mamba classifier, \( \mathbf{y} \) is passed through another linear layer to yield the final result of this module.

Finally, the output traverses the Pixel-Level Bi-Cross Attention module and a subsequent linear layer, culminating in the final binary classification result:
\begin{equation}
    \mathbf{y}_{cls}=\text{linear}(\text{Crossatten}(\text{mean}(\mathbf{y}),\mathbf{x}_M,\mathbf{x}_P))\in \mathbb{R}^2.
\end{equation}

\subsection{Pixel Level Bi-Cross Attention}
The classifier incorporates image features from MRI and PET scans along with tabular data during forward propagation. However, it struggles to effectively leverage pixel-level information from these images. Converting 3D MRI/PET directly into sequences results in long sequences, which hinder training efficiency. Additionally, the extensive image data can prevent the classifier from adequately integrating scale information. To tackle this, the Cross Attention architecture \cite{c5arenaza2018resistance} is proposed. This approach does not involve the classifier's forward propagation but instead employs attention mechanisms to make pixel-level information from MRI and PET available to the classifier's sequences.

As depicted in Fig. \ref{fig3} (Part B), the output from the final Mamba block in the classifier, denoted as \( \mathbf{y} \in \mathbb{R}^{1 \times d} \), is enhanced through mutual attention with MRI and PET before the final classification stage. The MRI and PET, initially represented as \( \mathbf{x}_M \) and \( \mathbf{x}_P \in \mathbb{R}^{H \times W \times D \times C} \), are reshaped into a summarized form \( \mathbf{x}_M \) and \( \mathbf{x}_P \in \mathbb{R}^{(H \cdot W \cdot D) \times C} \).

For MRI, the inter-attention process follows these steps:

\begin{gather}
    \mathbf{Q}_y=\mathbf{W}_q\mathbf{y},
    \mathbf{K}_x=\mathbf{W}_k\mathbf{x}_M,
    \mathbf{V}_x=\mathbf{W}_v\mathbf{x}_M,\\ 
    \mathbf{y}=\text{softmax}(\frac{\mathbf{Q}_y\mathbf{K}_x}{\sqrt{\mathbf{d}_k}})\mathbf{V}_x,
\end{gather}

where the query matrix (\(\mathbf{Q}_y\)) is obtained by linearly transforming the classifier's output, while the key (\(\mathbf{K}_x\)) and value (\(\mathbf{V}_x\)) matrices are derived from linearly transforming the serialized MRI features. The same mutual attention technique is applied to the PET.

After calculating mutual attention, these features are residually combined with \( \mathbf{y} \). Following feed-forward and layer normalization operations, they are once again residually summed with the original \( \mathbf{y} \).

\section{Experiment}
\subsection{Data Acquisition and Processing}
We validate our approach using the publicly available ADNI dataset. Our model training, as detailed in this paper, necessitates two distinct datasets: the MRI-PET dataset, comprising paired MRI and PET, and the MCI-AD dataset, used to evaluate the classifier's ability to predict progression to AD. Due to ADNI's privacy policy, we cannot publicly share the screened and processed datasets. Nonetheless, we will provide a comprehensive description of how these two datasets were constructed.

\textbf{MRI-PET Dataset.}
The dataset requirements are relatively flexible, necessitating corresponding MRI and PET scans from the same patient at the same diagnostic stage. This dataset must be substantial enough to support effective training on a generative network and representation learning. In our data collection process, we utilized the ADNI1, ADNI2, ADNI3, ADNI4, and ADNI-GO datasets. According to the literature \cite{c2jack2018nia, c1jia2014prevalence}, MRI and PET scans taken within ten days of each other accurately reflect the patient's condition at that time. For imaging protocols, we selected sagittal phase, 3D, T1-weighted MRI scans without preprocessing, specifically Magnetization Prepared RApid Gradient Echo. For PET scans, we chose 18F-FDG and applied preprocessing steps such as co-registration, averaging, standard normalization of image and voxel sizes, and uniform resolution adjustment. Our collection efforts resulted in 2,843 paired MRI and PET datasets, which we divided into 2,274 pairs for training and 569 pairs for validation. The 3D images, initially in DICOM format, were converted to NIfTI format.

\textbf{MCI-AD Dataset.}
To construct the dataset for this task, it was critical to ascertain each patient's status at each diagnostic stage. We used the tadpole table data from the ADNI study, which provides comprehensive basic and pathological information for each patient. Initially, we identified all patients diagnosed with MCI and then tracked their subsequent diagnoses over time. We documented the status and timing of each subsequent diagnosis, as shown in Fig. \ref{fig4}. If a patient was later diagnosed with AD, the classification label was set to 1; otherwise, it was set to 0.

\begin{table*}[]
\begin{center}
\caption{Ablation Experiments of GFE-Mamba on 1-year Dataset and 3-year Dataset (\%).}
\label{tab4}
\begin{adjustbox}{width=\textwidth, center} 
\begin{tabular}{@{}c|ccccc|ccccc@{}}
\hline
\multirow{2}{*}{Components}                                                 & \multicolumn{5}{c|}{1-Year Dataset}                        & \multicolumn{5}{c}{3-Year Dataset}                        \\
\cline {2-11}
& Precision($\uparrow$) & Recall($\uparrow$)    & F1-score($\uparrow$)  & Accuracy($\uparrow$)  & MCC($\uparrow$)       & Precision($\uparrow$) & Recall($\uparrow$)    & F1-score($\uparrow$)  & Accuracy($\uparrow$)  & MCC($\uparrow$)       \\
\hline
\begin{tabular}[c]{@{}c@{}}w/o Generative\\ Feature Extraction\end{tabular} & 88.57           & 83.33          & 89.29          & 86.21          & 76.60          & 86.21         & 65.00          & 92.86          & 76.47          & 69.28          \\
w/o Intermediate Features & 92.35 & 87.82 & 90.04 & 90.92 & 83.88 & 90.58 & 86.97 & 88.73 & 88.65 & 80.04 \\

w/o Bi-Cross Attention                                                      & 92.86          & 90.00          & 93.10          & 91.53          & 85.39          & 91.43          & 83.33          & 96.15          & 89.29          & 82.79          \\
w/o Vit Middle Block                                                        & 91.38          & 85.00          & 89.47          & 87.18          & 82.14          & 90.00          & 86.67          & 89.66          & 88.14          & 79.53          \\
w/o Image Data                                                              & 89.66          & 78.95          & 88.24          & 83.33          & 76.11          & 87.93          & 75.00          & 88.24          & 81.08          & 72.82          \\
w/o Table Data                                                              & 86.21          & 68.42          & 86.67          & 76.47          & 67.84          & 84.48          & 75.00          & 78.95          & 76.92          & 65.30          \\
\textbf{GFE-Mamba (Ours)}                                                    & \textbf{95.71} & \textbf{96.55} & \textbf{96.13} & \textbf{94.92} & \textbf{91.25} & \textbf{94.83} & \textbf{94.74} & \textbf{94.78} & \textbf{92.31} & \textbf{88.48} \\
\hline
\end{tabular}
\end{adjustbox}
\end{center}
\end{table*}

After constructing the classification labels, we identified the corresponding MRI using the information from our dataset. We retrieved the relevant MRI scans from the ADNI image dataset by matching the patient IDs and consultation dates recorded in the table. The MRI data, originally in DICOM format, was converted to NIfTI format for further analysis.

We then refined the tadpole table by adding a column to represent the time interval between diagnoses, denoted as \( \Delta t \), and removed extraneous information. This included: 1) redundant data, such as information for identity verification, 2) labeled information like diagnosis outcomes, and 3) complex metrics not pertinent to clinical diagnosis and training, such as specific brain region volumes. The average time interval between diagnoses was approximately 6.7 months, excluding extreme values (e.g., \( \Delta t \leq 90 \) days, or three months).

Following this processing, we created two datasets to evaluate methods across different time intervals: a one-year progression dataset and a three-year progression dataset. The one-year dataset includes 302 samples, divided into 242 for training and 60 for testing. The three-year dataset contains 351 samples, split into 281 for training and 70 for testing. The MCI-AD dataset comprises 136 positive samples and 155 negative samples. These datasets correspond to time intervals of \( 150 \leq \Delta t \leq 365 \) days and \( 150 \leq \Delta t \leq 1095 \) days, respectively.

Table \ref{tab1} presents the mean, variance, and counts of positive and negative samples for \( \Delta t \) in these datasets. To visualize the time intervals of all samples, Fig. \ref{fig7} displays box plots for the time intervals in the one-year and three-year progression datasets. The X and Y axes represent the dataset name and the time interval length, respectively. Each dataset is represented by two box plots: one before and one after excluding outliers. After outlier removal, the distribution of time intervals is more concentrated and extends over a larger range, enhancing the model's ability to learn over extended periods.

\subsection{Evaluation Indicators}
\label{subsec:Evaluation Indicators}
We utilized five metrics to assess the classification performance on the dataset obtained from ADNI: Accuracy, Precision, Recall, F1-score, and Matthews Correlation Coefficient (MCC). Below are their definitions and formulas:
\begin{gather}
    \mathrm{Precision}=\frac{TP}{TP+FP}, \\
    \mathrm{Recall}=\frac{TP}{TP+FN}, \\
    \text{F1-score}=2\times \frac{Precision\times Recall}{Precision + Recall}, \\
    \mathrm{Accuracy}=\frac{TP+TN}{TP+TN+FP+FN}, \\
    \mathrm{MCC}=\frac{TP\times TN - FP\times FN}{\sqrt{(TP+FP)(TP+FN)(TN+FP)(TN+FN)}},
\end{gather}

    
where \(TP\) (True Positives) and \(TN\) (True Negatives) indicate the correctly predicted positive and negative samples, respectively. Conversely, \(FP\) (False Positives) and \(FN\) (False Negatives) denote the incorrectly predicted positive and negative samples.
 
\begin{figure}
\centering
    \includegraphics[width=\linewidth]{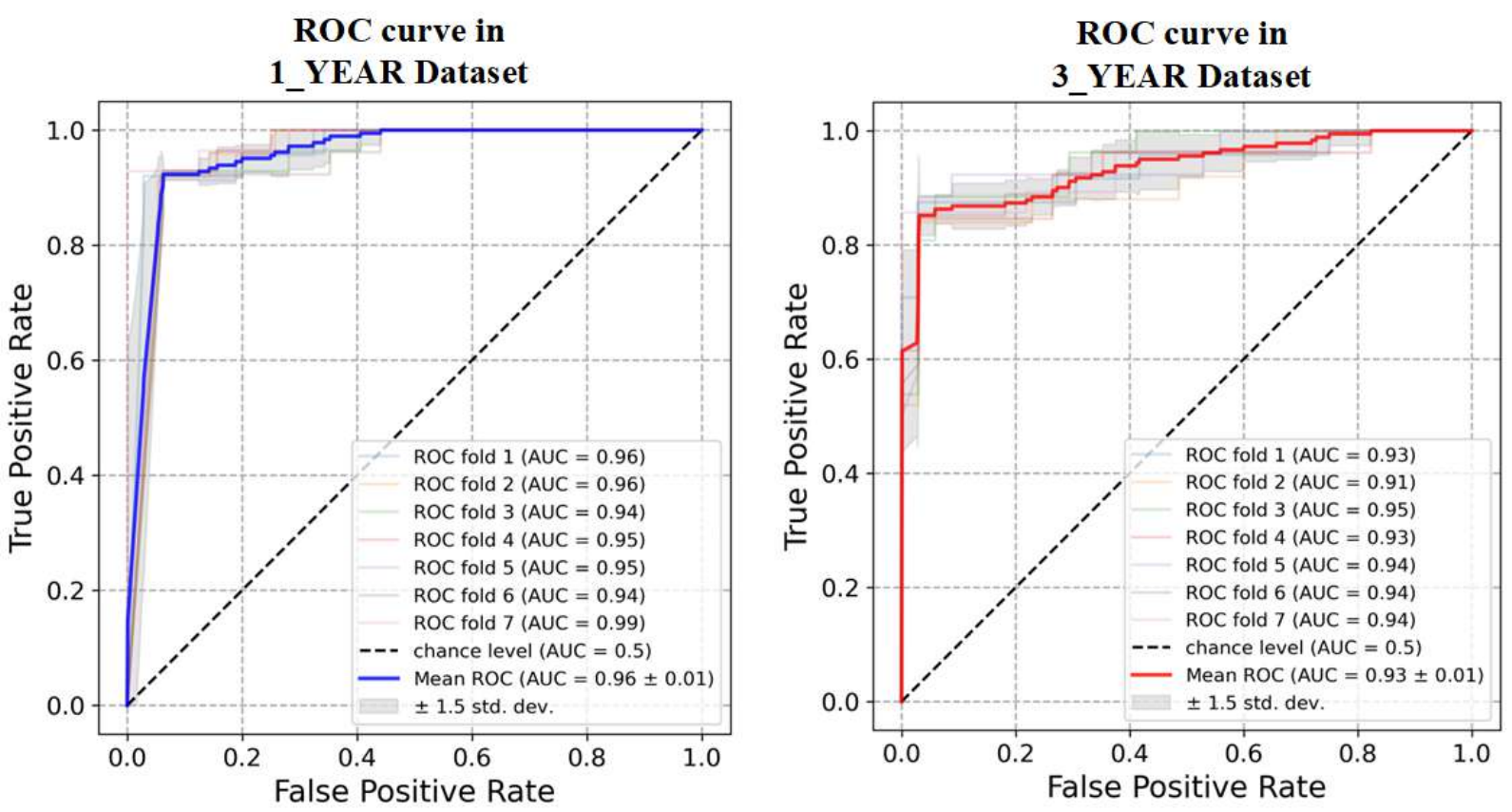}
    \caption{The ROC curve that was generated through 7-fold cross-validation on the one-year and three-year datasets. The semi-transparent lines represent the ROC from each of the 7 cross-validations, while the opaque line indicates the average. The shaded gray area represents the error level within one and a half standard deviation.}
    \label{fig9}
\end{figure}

\subsection{Experimental Settings}
Both components of the study were implemented using PyTorch 2.0 on NVIDIA GeForce RTX 4090 GPUs with CUDA 11.8. We used Monai to read images in NIfTI format and Pandas to read tables and convert them into training data. The 3D GAN-ViT model was trained over 200 epochs with a batch size of 2, while the classifier was trained for 100 epochs with a batch size of 8. Both models were optimized using the Adam algorithm, with a learning rate of 0.0001 and beta parameters set at (0.9, 0.999). The experiment consists of comparative and ablation studies: In the comparative study, both JSRL and our method first train the GAN network on the aforementioned MRI-PET Dataset. Subsequently, all methods are trained on the MCI-AD Dataset. In the ablation study, all other parameters are kept constant while sequentially removing GFE, Intermediate Feature, Pixel Level Bi-Cross Attention, Vit Middle Block, Image Data, and Table Data. After training each group, the various metrics in \ref{subsec:Evaluation Indicators} are measured.

\subsection{Comparative Experiments}
The experimental results comparing our GFE-Mamba model with other advanced AD classification models using the ADNI dataset are presented in Table \ref{tab2}. We assessed our method against unimodal approaches, which utilize either CT scans or assessment scales independently, and multimodal approaches, which integrate both CT scans and assessment scales. For the unimodal comparison using CT scans, we selected 3D ResNet \cite{c27he2016deep}, PE-Net \cite{huang2020penet}, and JSRL \cite{liu2022assessing}. PE-Net is noted for its significant results in lung CT diagnosis using Squeeze-and-Excitation blocks, while JSRL excels in diagnosing brain CT through representation learning. For unimodal assessment scale approaches, we considered TabTransformer \cite{c28huang2020tabtransformer}, FTTransformer \cite{gorishniy2021revisiting}, and XGBoost \cite{c29kavitha2022early}. XGBoost stands out for its performance in tabular tasks using the Gradient Boosting Decision Tree, and both TabTransformer and FTTransformer excel with long tables through the use of transformers. In the multimodal category, we evaluated methods from Radfusion \cite{zhou2021radfusion}, the Fusion model \cite{c31qiu2022multimodal}, as well as approaches by Qiu \textit{et al.} and \textit{Zhang et al.}, all of which accurately diagnose using both CT images and textual information. The results indicate that GFE-Mamba significantly outperforms these models, particularly in terms of MCC and Accuracy. Due to the consistency and robustness of our findings across datasets, we focus our analysis on the comparative results from the 1-year dataset.

\begin{figure}
\centering
    \includegraphics[width=\linewidth]{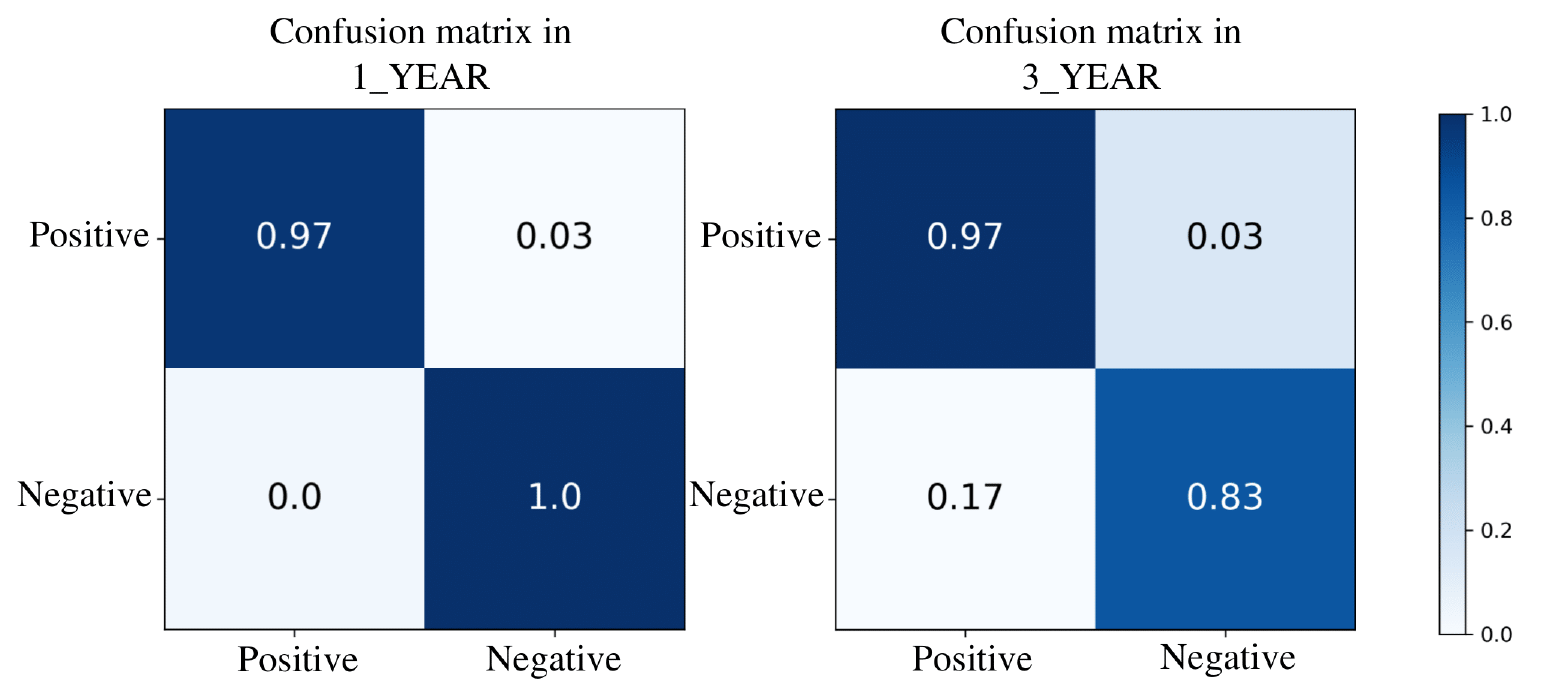}
    \caption{The confusion matrix for GFE-mamba. The values in each row have been normalized to assess the classification performance for both positive and negative cases.}
    \label{fig10}
\end{figure}

Compared to the ResNet family of models \cite{c27he2016deep}, which are designed to address gradient vanishing issues in deep networks, GFE-Mamba demonstrates superior performance in handling MRI. Specifically, the ResNet50 model falls short in Precision and Accuracy, achieving only 81.03\% and 73.17\%, respectively. While ResNet models often struggle to capture localized pathology, GFE-Mamba effectively processes spatial vectors using the 3D GAN-ViT module, enhancing the capture of spatial information and improving classification accuracy. 

Similarly, when compared to TabTransformer \cite{c28huang2020tabtransformer}, known for its strength in tabular data processing, GFE-Mamba shows a more robust ability to capture pathological features in MRI and recognize complex pathological states. The TabTransformer model records lower Recall and F1-score, at 90.57\% and 87.32\%, respectively. The results illustrate that the integration of the 3D GAN-ViT module and the Multimodal Mamba Classifier in GFE-Mamba significantly boosts classification accuracy and model interpretability. 

When evaluated against traditional AD classification models like XGBoost \cite{c29kavitha2022early} and Qiu \textit{et al.}'s model \cite{c30qiu2020development}, GFE-Mamba addresses the limitations of feature extraction capability and parameter redundancy inherent in these models, which rely heavily on traditional CNNs. Consequently, GFE-Mamba outperforms both the Early-Stage model and the 3D CNN model in Recall and F1-score, which achieved 86.53\% and 86.92\%, and 81.82\% and 84.17\%, respectively. This means that GFE-Mamba overcomes the high computational complexity and inadequate global information capture typical of 3D CNN models by incorporating a Pixel-Level Bi-Cross Attention mechanism, thereby enhancing feature expression capability and model interpretability. 

Furthermore, compared to advanced AD classification models such as the Fusion model \cite{c31qiu2022multimodal} and Zhang \textit{et al.}'s model \cite{c32zhang2023alzheimer}, GFE-Mamba showcases superior performance. While these models perform well in multimodal data processing and feature extraction, they often encounter issues with feature redundancy and nonlinear feature representation, particularly in complex neuroimaging data. GFE-Mamba addresses these challenges by combining the 3D GAN-ViT module with the Multimodal Mamba Classifier, thereby minimizing spatial and channel redundancy and optimizing feature representation. The Pixel-Level Bi-Cross Attention mechanism further improves nonlinear feature representation and model interpretability, while reducing memory usage and computational complexity. As a result, GFE-Mamba excels in capturing MRI features and accurately distinguishing complex pathological states, outperforming the Multimodal deep learning model and other AD classification models, which recorded lower Precision and F1-score of 89.83\%, 88.91\%, and 76.67\%, 79.86\%, respectively.

To thoroughly assess the effectiveness of our method, we conducted a 7-fold cross-validation on both the one-year and three-year datasets. For each round of validation, we plotted the ROC curves and also included the average ROC curve to illustrate overall performance. As depicted in Fig. \ref{fig9}, GFE-Mamba consistently shows high performance in classification tasks, maintaining stability across different datasets. In addition to the ROC analysis, we gathered the model's predicted outcomes and calculated the average counts of true positives, false positives, true negatives, and false negatives. Using these metrics, we constructed a confusion matrix. As illustrated in Fig. \ref{fig10}, our method demonstrates high accuracy in correctly classifying both positive and negative cases.

\subsection{Ablation Study}
In the ablation experiments section, we examine the distinct contributions of the GFE, Pixel-Level Bi-Cross Attention, and ViT middle block components to the classification performance of the GFE-Mamba model on both the 1-year and 3-year datasets. We assessed the impact of each module by analyzing changes in Accuracy, Precision, Recall, F1-score, and MCC values when each component was individually removed, compared to the complete GFE-Mamba model. The results, as shown in Table \ref{tab4}, illustrate that each module significantly enhances the model's classification performance. Due to the consistent results across both datasets, our discussion will focus on the ablation experiments conducted on the 1-year dataset.

\begin{figure*}
\centering
    \includegraphics[width=\textwidth]{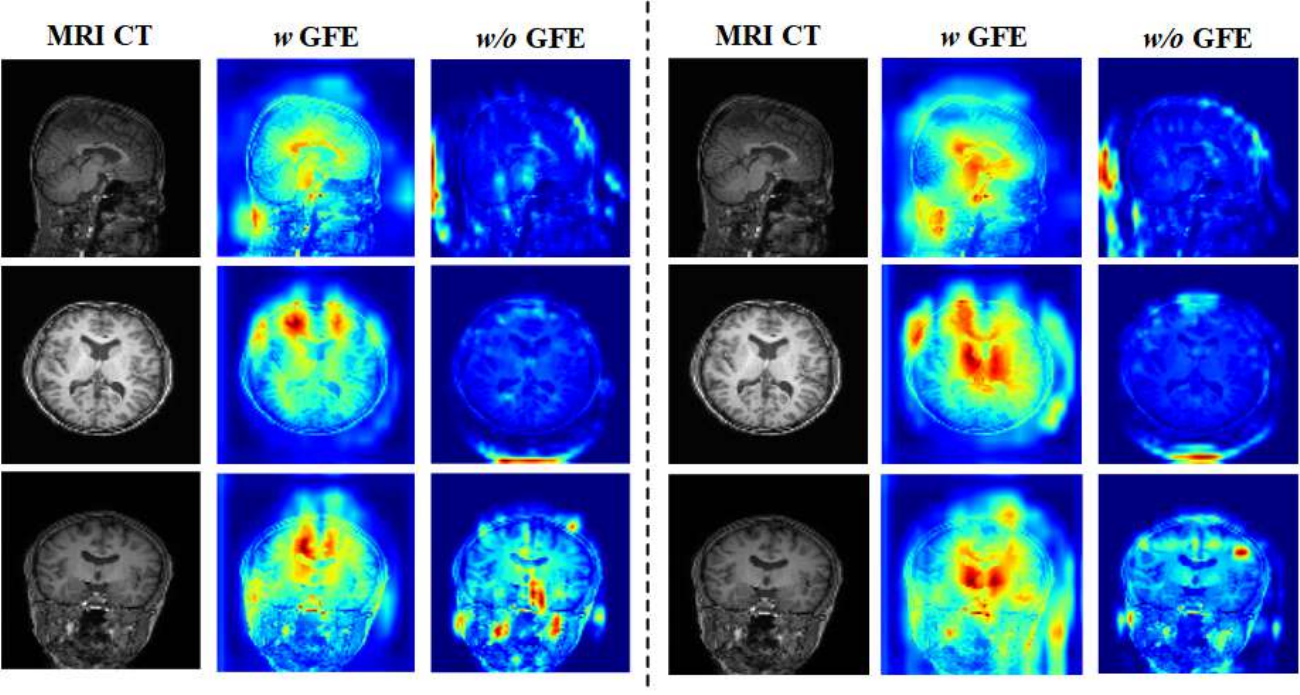}
    \caption{The heatmaps of two MRI CT images by GradCAM during classification. The first, second, and third rows represent the sagittal, axial, and coronal planes, respectively. GFE represents Generative Feature Extractor.}
    \label{fig8}
\end{figure*}

\textbf{Impact of Removing Generative Feature Extraction.} The GFE module is crucial for enhancing the model's ability to extract features from high-dimensional neuroimaging data by utilizing GANs. When this module is removed, the 3D GAN-ViT is not trained for generative tasks and retains only its encoder. This removal considerably hampers the model's feature extraction capabilities, leading to a significant drop in performance. For instance, Precision falls from 95.71\% to 88.57\%, and the F1-score declines from 96.13\% to 89.29\%. Other metrics also declined. These results highlight the essential role of the GFE module in optimizing feature representation, particularly in capturing detailed MRI features. To further illustrate the effectiveness of this module, we present a comparison using GradCAM in Fig. \ref{fig8}, showcasing our method with and without the Generative Feature Extractor. GradCAM helps explain which parts of the image the model focuses on by utilizing the weights from the global average feature map of the gradients. For visualizing the 3D MRI and CT images, the CT and its corresponding feature maps are sliced from the sagittal, axial, and coronal anatomical planes with a thickness of half slices. The visualization shows that with the Generative Feature Extractor, the model concentrates more on relevant brain regions. In contrast, without it, the model lacks clearly focused areas.

\textbf{Impact of removing Intermediate Features.} The intermediate features from the GAN module are crucial for feeding into the classifier because they provide a more comprehensive feature set than those typically used solely for classification. This is due to the encoder trained via the GAN extracting a broader range of features. Moreover, the generated PET images rely on these extracted intermediate features. To illustrate the importance of these intermediate features in generating PET images from MRIs, we removed them during the PET generation process. Instead, we used an encoder made of 3D convolutional and pooling layers to extract features from the generated PET, training this encoder alongside the classifier. The removal of intermediate features led to a decrease in classifier performance, with precision, recall, F1 score, accuracy, and MCC dropping by 3.36\%, 8.73\%, 6.09\%, 4.00\%, and 7.37\%, respectively. These results demonstrate that directly extracting intermediate features during the PET generation process is more beneficial for classifier discrimination than extracting features from the generated PET.

\textbf{Impact of Removing Bi-Cross Attention Module.} The Pixel Level Bi-Cross Attention module plays a vital role in enhancing the model’s ability to identify relationships between different data modalities, thereby improving feature representation and interpretability. Removing this module significantly diminishes the model's capacity to integrate multimodal data, resulting in a notable decline in performance. Specifically, Recall drops by 6.55\%, and the MCC decreases by 5.86\%. These findings underscore the critical importance of this mechanism in extracting and integrating information from multimodal data for a comprehensive understanding and accurate classification of complex pathological features.

\textbf{Impact of Removing ViT Middle Block.} The ViT middle block is essential for enhancing the model's ability to capture global spatial information, enabling it to manage diverse spatial relationships and subtle features in MRI. Replacing this block with a residual block reduces the model’s capacity to capture these global spatial features, resulting in decreased performance. Notably, Accuracy falls by 7.74\%, and Recall drops from 11.55\%. This reduced ability to extract global features complicates the model's task of distinguishing complex pathologies, highlighting the ViT middle block's critical role in identifying intricate pathological states.

\textbf{Impact of Removing Image Data.} Image data is fundamental to the model's performance. Its absence significantly diminishes the model's feature extraction capabilities and accuracy in recognizing pathological conditions. Without visual cues, the model's classification efficiency is impaired, leading to a decline in performance. Specifically, Precision decreases from 6.05\%, and MCC falls from 15.14\%. These results emphasize the importance of image data in capturing essential pathological features for accurate diagnosis.

\textbf{Impact of Removing Tabular Data.} Tabular data is crucial for successful multimodal data fusion. Removing it weakens the model's ability to integrate information from diverse sources, impairing its comprehensive understanding of pathological features. The absence of tabular data limits the model's effectiveness in utilizing multi-source information, causing a significant drop in performance metrics. Specifically, Accuracy decreases from 18.45\%, and the F1-score declines from 9.46\%. These findings underscore the importance of tabular data in complementing image data for accurate and effective classification.

\section{Conclusion}

The GFE-Mamba model proposed in this paper addresses challenges in multimodal data fusion, feature expressiveness, and model interpretability in predicting the progression from MCI to AD. By integrating the 3D GAN-ViT, Multimodal Mamba Classifier, and Pixel-Level Bi-Cross Attention mechanism, the GFE-Mamba model effectively extracts pathological features from MRI, utilizing scale information for robust fusion and classification prediction. This capability ensures strong performance in AD classification tasks, even with incomplete data from the ADNI Dataset.

Given the open-access limitations of the ADNI dataset, we have outlined the processes for collecting, processing, and making predictions from public datasets like ADNI. We provide the associated processing code to support other researchers in conducting spatiotemporal predictions. Through comparative and ablation experiments on reconstructed one- and three-year ADNI prediction datasets, we validate the effectiveness and resilience of the GFE-Mamba model. It significantly surpasses other state-of-the-art models in terms of Accuracy, Precision, Recall, F1-score, and MCC. Our ablation experiments reveal that removing any module component results in a notable decline in performance, underscoring the crucial role each part plays in fully extracting and utilizing multimodal data to enhance classification accuracy.

Looking ahead, as the field advances, we plan to collaborate with partner hospitals to gather larger and more diverse datasets. This includes extending the average size of time intervals in the dataset and collecting data from subjective cognitive decline through to MCI datasets, thus improving the model's generalization and robustness. Ultimately, we aim to develop an integrated system that allows patients to quickly receive categorical judgments and progression predictions from a single MRI and scale, thereby reducing the diagnostic burden on clinicians.


\end{document}